\documentclass{article}

\usepackage{microtype}
\usepackage{graphicx}
\usepackage{subfigure}
\usepackage{booktabs} 

\usepackage{hyperref}

\usepackage[accepted]{tom2023}

\usepackage{amsmath}
\usepackage{amssymb}
\usepackage{mathtools}
\usepackage{amsthm}

\usepackage[capitalize,noabbrev]{cleveref}

\theoremstyle{plain}

\theoremstyle{definition}

\theoremstyle{remark}

\usepackage[textsize=tiny]{todonotes}

\begin{document}

\twocolumn[
\icmltitle{Theory of Mind as Intrinsic Motivation for Multi-Agent Reinforcement Learning}



\icmlsetsymbol{equal}{*}

\begin{icmlauthorlist}
\icmlauthor{Ini Oguntola}{cmu}
\icmlauthor{Joseph Campbell}{cmu}
\icmlauthor{Simon Stepputtis}{cmu}
\icmlauthor{Katia Sycara}{cmu}
\end{icmlauthorlist}

\icmlaffiliation{cmu}{School of Computer Science, Carnegie Mellon University, Pittsburgh, USA}

\icmlcorrespondingauthor{Ini Oguntola}{ioguntol@andrew.cmu.edu}

\icmlkeywords{Theory of Mind, ToM, RL, multi-agent, reinforcement learning, intrinsic motivation}

\vskip 0.3in
]



\printAffiliationsAndNotice{}  

\begin{abstract}
The ability to model the mental states of others is crucial to human social intelligence, and can offer similar benefits to artificial agents with respect to the social dynamics induced in multi-agent settings. We present a method of grounding semantically meaningful, human-interpretable beliefs within policies modeled by deep networks. We then consider the task of \textit{2nd-order} belief prediction. We propose that ability of each agent to predict the beliefs of the other agents can be used as an intrinsic reward signal for multi-agent reinforcement learning. Finally, we present preliminary empirical results in a mixed cooperative-competitive environment.
\end{abstract}

\section{Introduction}

The ability to infer the mental states of oneself and others -- beliefs, desires, intentions, preferences, etc -- is known as \textit{theory of mind} (ToM) \cite{baker2011bayesian}. Humans naturally build rich internal models of others, and are able to use these inferences to predict the behavior of others, to condition their own behavior, and to forecast social interactions \cite{georgeff1999belief}. Theory of mind has long been studied within cognitive science and psychology \cite{premack1978does}, a fundamental aspect of human social intelligence that has been shown to develop in early childhood. \cite{ensink2010development, astington2010development}.

Traditionally, agent-modeling approaches within reinforcement learning (RL) and imitation learning largely ignore the idea of internal mental states, typically only focused on modeling external actions \cite{he2016opponent, wen2019probabilistic}. However, there is a growing body of work in the machine learning literature aimed towards developing artificial agents that exhibit theory of mind  \cite{baker2011bayesian, rabinowitz2018machine, jara2019theory, fuchs2021theory}. Even beyond simply providing a helpful inductive bias for modeling behavior, ToM reasoning has the potential to enable the discovery and correction of false beliefs or incomplete knowledge, facilitate efficient communication and coordination, and improve human-agent teaming \cite{zeng2020brain, sclar2022symmetric, oguntola2021deep}.

The work of \cite{aru2023mind} highlights key challenges regarding the difficulty of evaluating current deep learning ToM approaches. In particular, from a human perspective we may solve a task using an already-developed internal theory of mind, whereas an artificial agent may be able to learn simpler decision rules or take advantage of spurious correlations as shortcuts, and it is difficult to determine whether ToM has actually been learnt.

Here we consider the reverse -- rather than solving a task and hoping it induces a theory of mind, we instead explicitly learn a theory of mind over semantically grounded beliefs, and use this as a signal to solve the task. Our fundamental research question is the following: can modeling other agents' \textit{beliefs} serve as an intrinsic reward signal to improve performance in multi-agent settings?

In this paper we develop an approach to explicitly grounding semantically meaningful beliefs within RL policies. We then propose the use of ToM reasoning over the beliefs of other agents as intrinsic motivation in multi-agent scenarios. We run experiments in a mixed cooperative-competitive environment and show preliminary results that suggest this approach may improve multi-agent performance, with respect to both coordination and deception.

The primary contributions of this paper are the following:
\begin{itemize}
    \item We develop an information-theoretic residual variant to the concept bottleneck learning paradigm \cite{koh2020concept} based on mutual information minimization.
    \item We utilize this approach to model semantically-meaningful belief states within RL policies.
    \item We propose the prediction task of second-order prediction of these beliefs (i.e. ToM reasoning) as intrinsic motivation.
    \item We demonstrate preliminary results that demonstrate improved performance in a mixed cooperative-competitive environment.
\end{itemize}

\section{Related Work}

\subsection{Intrinsic Motivation in Deep RL}

Intrinsic motivation in reinforcement learning refers to the use of an additional reward signal to encourage particular agent behaviors without direct feedback from the environment on the task.

In the single-agent setting, common approaches to intrinsic motivation include ``curiosity" to encourage visiting novel states \cite{pathak2017curiosity} and ``empowerment" to encourage diversity of reachable states \cite{mohamed2015variational}.

Most of these approaches can also be extended to the multi-agent setting, but the introduction of multiple agents inherently creates an inter-agent dynamic that can be explored as well. \cite{jaques2019social} proposed an intrinsic reward for ``social influence" by rewarding agents for having high mutual information between their actions. \cite{wang2020influence} develop similar approaches that reward an agent for influencing the state transition dynamics and rewards of other agents.

In constrast, our intrinsic reward approach is predicated on influencing the internal beliefs of other agents, rather than directly influencing their external states or actions.

\subsection{Theory of Mind in Multi-Agent RL}

Although RL often implicitly involves theory of mind via agent modeling, recent approaches have also sought to model this directly \cite{rabinowitz2018machine}.

Within multi-agent reinforcement learning there have been a variety of approaches inspired by ToM reasoning, modeling beliefs \cite{fuchs2021theory, wangtom2c, sclar2022symmetric} and intents \cite{qi2018intent, xu2019intentmeta}. Other inverse reinforcement learning methods approach ToM-like reasoning by conditioning the reward function on inferred latent characteristics \cite{tian2021learning, wu2023multiagent}. Most of these are aimed at improving coordination in cooperative multi-agent scenarios, particularly with regard to communication \cite{sclar2022symmetric, wangtom2c}.

\subsection{Concept Learning}

Concept learning, generally speaking, is an approach to interpretability for deep neural networks that involves enforcing structure on the latent space to represent grounded, semantically meaningful ``concepts".

One such approach is concept whitening \cite{chen2020concept}, in which an intermediate layer is inserted for orthogonal alignment of data in the latent space with predefined human-interpretable concept labels, with concepts provided via auxiliary datasets. The restriction with this method is the inherent assumption that all concepts are non-overlapping.

Concept bottleneck models are a similar approach developed an approach that consists of a concept extractor directly supervised on concept labels, and a predictor network that generates an output from these concepts \cite{koh2020concept}. While more flexible than concept whitening in the sense that it can encode any set of concepts, it still makes the assumption that the provided set of concepts alone is expressive enough for the predictive task; performance suffers when this not the case.

Some approaches mitigate this by combining the concept predictions with a residual extracted from the input, they either impose additional constraints (e.g. orthogonality) on the combined output that may not hold \cite{renos}, or they do not provide a way to directly ensure the information encoded by the residual does not overlap with the concepts \cite{yuksekgonul2022post}, allowing the model to effectively ignore concepts in its decision making process.

While prior work has used these approaches in the context of imitation and reinforcement learning \cite{oguntola2021deep, renos}, in this work we specifically examine concept learning as a way to approach the challenge of grounding semantically meaningful \textit{mental states} within policies. We also develop a residual variant that directly encourages decorrelation between concepts and residual while avoiding the introduction of any restrictive assumptions.





\section{Method}

\begin{figure*}[ht]
\vskip 0.2in
\begin{center}
\centerline{\includegraphics[width=0.85\linewidth]{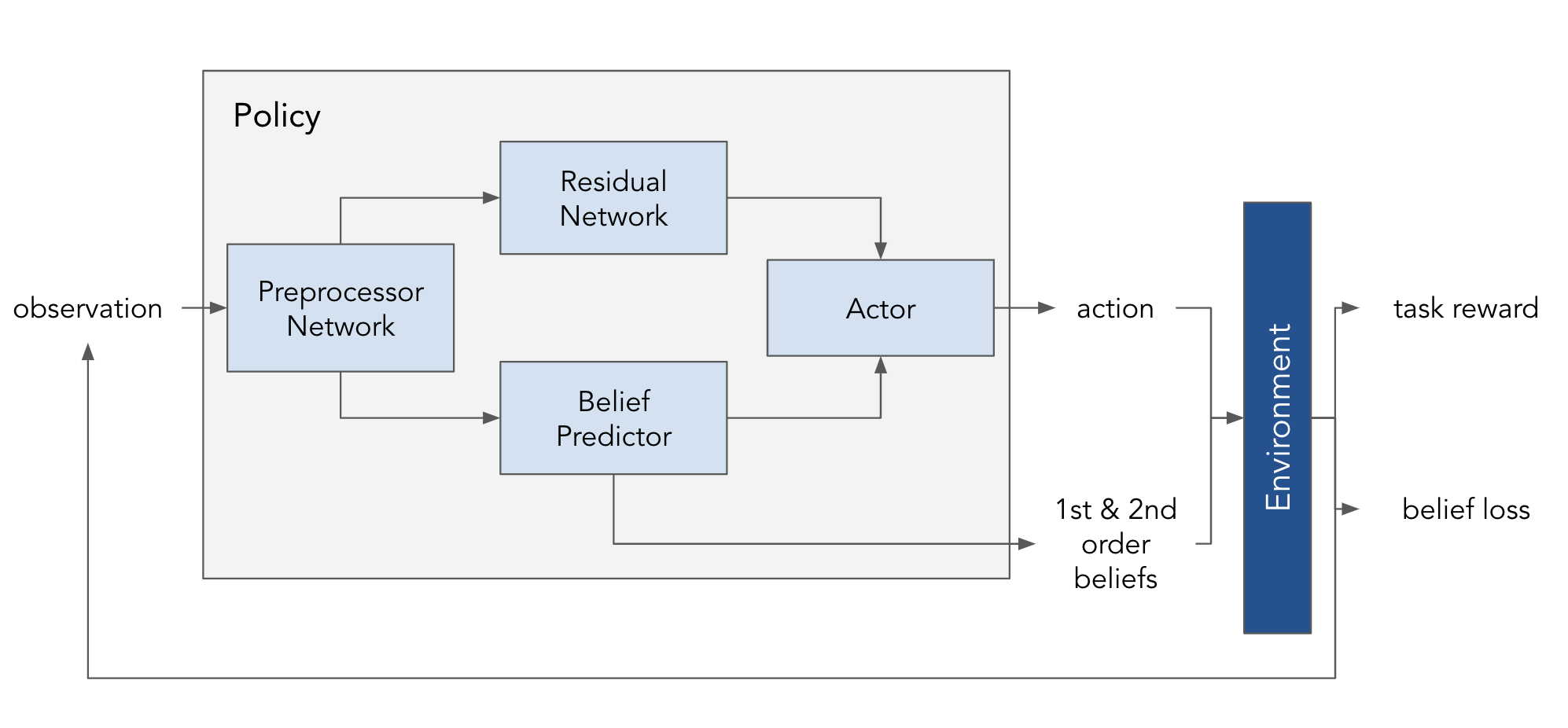}}
\caption{Policy models with 1st and 2nd-order belief prediction. The belief predictor is supervised by ground truth labels, and the residual network is regularized via mutual information minimization with respect to beliefs.}
\label{fig:policy}
\end{center}
\vskip -0.2in
\end{figure*}

\subsection{Modeling Beliefs via Concept Learning}

In deep reinforcement learning, policies are typically black box models that directly map states to actions. Our approach follows the paradigm of concept learning \cite{yi2018neural, chen2020concept, koh2020concept, yeh2020completeness, oguntola2021deep, renos}, which involves inserting an intermediate \textit{concept layer} which is designed to align with human-interpretable ``concepts", typically via a supervised auxiliary loss. In our setting, these concepts are designed to model \textit{beliefs} about the environment. For instance, in an environment with a door, one could model the belief over whether the door is locked as a binary concept $b_{locked} \in \{0, 1\}$.
\begin{align}
    L_{belief} = 
        \begin{cases}
            \mathrm{MSE}(\mathbf{b}, \mathbf{b}') & \text{if continuous} \\
            \mathrm{CE}(\mathbf{b}, \mathbf{b}') & \text{if discrete}
        \end{cases}
\end{align}
where $\mathbf{b}$ is the agent belief vector, $\mathbf{b}'$ is the ground truth, MSE is the mean-squared error, and CE is the cross entropy loss. 

These beliefs are then used to generate an action. However, depending on the selection of beliefs, they alone may not be a sufficient signal to learn a policy that successfully solves a given task. We mitigate this by additionally introducing a \textit{residual} -- a compressed representation of the input that is concatenated to the belief vector. Given vector input $\mathbf{x}$, we have our residual network generate $r(\mathbf{x}) = \mathbf{z}$.

It is important that our residual and beliefs be disentangled -- that is, the residual should not contain any information about the beliefs -- as otherwise our model may simply learn to rely entirely on the residual and ignore the beliefs, which would compromise the interpretablity of the policy.

We approach ``disentanglement" from a probability theory perspective, aiming to ensure that the belief and residual vectors are statistically independent. Here our goal is to minimize the mutual information between the belief vector and residual, which is zero if and only if they are independent. This measure can also be characterized as KL-divergence between the joint distribution and the product of the marginal distributions:
\begin{align}
    I(B;Z) = D_{KL}(\mathbb{P}_{BZ} \parallel \mathbb{P}_B \otimes \mathbb{P}_Z)
\end{align}

To achieve this, we utilize the variational approach from \cite{club} and minimize a contrastive log-ratio upper bound:
\begin{align}
    L_q(\theta) &= -\mathbb{E}_{p_\sigma(\mathbf{b}, \mathbf{z})}[\log q_\theta(\mathbf{z} | \mathbf{b})] \\
    \nonumber L_{residual}(\sigma) &= \mathbb{E}_{p_\sigma(\mathbf{b}, \mathbf{z})} [\log q_\theta(\mathbf{z} | \mathbf{b})] \\
    & \quad\quad - \mathbb{E}_{p_\sigma(\mathbf{b})} \mathbb{E}_{p_\sigma(\mathbf{z})} [\log q_\theta(\mathbf{z} | \mathbf{b})]
\end{align}
where $\mathbf{b}$ is the belief vector, $\mathbf{z}$ is the residual vector, $p_\sigma(\mathbf{b}, \mathbf{z})$ is the joint distribution of intermediate outputs from our policy, and $q_\theta(\mathbf{z} | \mathbf{b})$ is a variational approximation to the conditional distribution $p_\sigma(\mathbf{z} | \mathbf{b})$, modeled via a separate neural network trained to minimize negative log-likelihood $L_q(\theta) = -\log\mathcal{L}(\theta)$.

Unlike approaches based on concept whitening \cite{oguntola2021deep, renos}, our method of disentanglement does not assume or impose any intra-dimensional orthogonality constraints within the concept (i.e. belief) or residual layers, but rather decorrelates the two vectors as a whole. Specifically, we make no restrictive assumptions that concepts are mutually exclusive, and also retain full multi-dimensional expressiveness within our residual representation while simultaneously minimizing correlation with our concept vector.

Finally, the concatenated output $(\mathbf{b}, \mathbf{z})$ is fed into the rest of the actor network to generate an action. The concept layer and residual layer are trained by adding the additional loss terms to the objective function optimized by the reinforcement learning algorithm of choice. For our experiments we use the PPO objective from \cite{ppo}, but generally speaking this approach is agnostic to the particular RL algorithm chosen.
\begin{align}
    L_{PPO}(\sigma) &= \mathbb{E}_t [ \min ( r_t(\sigma)A_t, \\
    & \quad\quad\quad\quad\quad \nonumber \mathrm{clip}(r_t(\sigma), 1 + \epsilon, 1 + \epsilon) A_t ) ] \\
    L_{policy} &= \alpha L_{PPO} + \beta L_{belief} + \gamma L_{residual}
\end{align}
where $r_t(\sigma) = \frac{\pi_{\sigma}(a_t | s_t)}{\pi_{\sigma_{old}(a_t | s_t)}}$ is the PPO probability ratio, $\pi_\sigma$ is the policy to be optimized, $A_t$ is the advantage function, and $\alpha, \beta, \gamma, \epsilon > 0$ are hyperparameters.

During training, for each batch we optimize both the policy loss $L_{policy}$ (with respect to the policy parameters $\sigma$) and the variational loss $L_q$ (with respect to the variational parameters $\theta$).


\subsection{Second-Order Belief Prediction}

In a multi-agent scenario where each agent is reasoning over the same set of beliefs over the environment, consider the \textit{second-order belief} as one agent's prediction of another agent's beliefs. It is important to note that the first-order belief of an agent may be incorrect, in which case a correct second-order belief would successfully predict this false belief.

For instance, consider a scenario where a door is locked but agent A believes the door is unlocked. Agent B should ideally have 1) the first-order belief that the door is unlocked, and 2) the second-order belief that agent A thinks the door is locked.

Our approach proposes the use of second-order belief prediction as an intrinsic reward. Intuitively speaking, we want to incentivize each agent to 1) learn to predict the beliefs of other agents and 2) learn to behave in a way such that the beliefs of the other agents will be predictable (e.g. learning to observe other agents, learning to communicate, etc).

We do this by augmenting the agent's belief network to produce not only its own belief vector, but also a belief vector prediction for each of the other agents.
\begin{align}
    \mathbf{B} = \left[ \mathbf{b} + f(\mathbf{x})_i \right]_{i=1}^K
\end{align}
where $K$ is the total number of agents, $\mathbf{B}$ is the $K \times dim(\mathbf{b})$ second-order belief matrix, and $f : \mathbb{R}^{\mathrm{dim}(\mathbf{x})} \rightarrow \mathbb{R}^{K \times \mathrm{dim}(\mathbf{b})}$ is modeled by a neural network.

Rather than treat this as a directly-supervised auxiliary task, we instead include the second-order prediction loss as an additional reward term, as we want the policy's value estimation to be biased towards states where both the current and the \textbf{future} beliefs (or belief distributions) of the other agents tend to be predictable (e.g. states where it can gain information about other agents).

Then the intrinsic reward becomes the negative belief prediction loss:
\begin{align}
    r_{tom} &= 
        \begin{cases}
            - \frac{1}{K}\sum_{i=1}^K MSE(\mathbf{B}_i, \mathbf{b}^{(i)}) & \text{if continuous} \\
            - \frac{1}{K}\sum_{i=1}^K CE(\mathbf{B}_i, \mathbf{b}^{(i)}) & \text{if discrete}
        \end{cases} \\
    r &= r_{task} + \lambda r_{tom}
\end{align}
where $\lambda \geq 0$ is a hyperparameter.

\subsection{Training vs Execution}

The training setup requires that all agents are trained in the manner previously described, and we assume that the beliefs of other agents are available during centralized training to calculate intrinsic reward.

During training we do not propagate gradients from the policy or reward through the 1st-order belief prediction network; that is, the 1st-order belief prediction network is only updated from the supervised belief loss on ground truth values from the environment, and is unaffected by the reward dynamics of the task. In combination with the mutual information regularization for the residual, this ensures that any belief information relevant to an agent's policy comes only from the agent's ability to infer the correct values of said beliefs from the environment. This approach eliminates any potential issues with a "malicious actor" purposefully generating incorrect belief predictions.

Execution, on the other hand, does not require beliefs or any inner states of other agents, and thus can be done with other policies that were not trained with our training setup or architecture -- or even with human agents.




\section{Experiments}

\begin{table*}[htb]
\caption{Performance on ParticleWorld physical deception task, in various configurations. Here we present the mean cumulative reward of the final trained policies, averaged across 5 random seeds, where (good) is used to indicate the green good agents, and (adv.) is used to indicate the red adversary. With respect to beliefs, we vary whether the each policy generates 1st-order predictions, 2nd-order predictions, or none at all. Episode reward variance is given in parentheses.}
\label{adversary-rewards}
\vskip 0.15in
\begin{center}
\begin{small}
\begin{sc}
\begin{tabular}{cccccc}
\toprule
1st-Order & 1st-Order & 2nd-Order & 2nd-Order & Episode Reward & Episode Reward \\
(Good) & (Adv.) & (Good) & (Adv.) & (Good) & (Adv.) \\
\midrule
No & No & No & No & 1.889 ($\pm$ 0.23) & -15.32 ($\pm$ 0.51) \\
Yes & Yes & No & No & 2.209 ($\pm$ 0.11) & -15.17 ($\pm$ 0.29) \\
Yes & Yes & \textbf{Yes} & No & \textbf{2.760 ($\pm$ 0.44)} & -17.78 ($\pm$ 0.32) \\
Yes & Yes & No & \textbf{Yes} & 1.636 ($\pm$ 0.41) & \textbf{-14.01 ($\pm$ 0.30)} \\
\bottomrule
\end{tabular}
\end{sc}
\end{small}
\end{center}
\vskip -0.1in
\end{table*}


\subsection{ParticleWorld: Physical Deception}

\begin{figure}[htbp]
\vskip 0.2in
\begin{center}
\centerline{\includegraphics[width=0.6\linewidth]{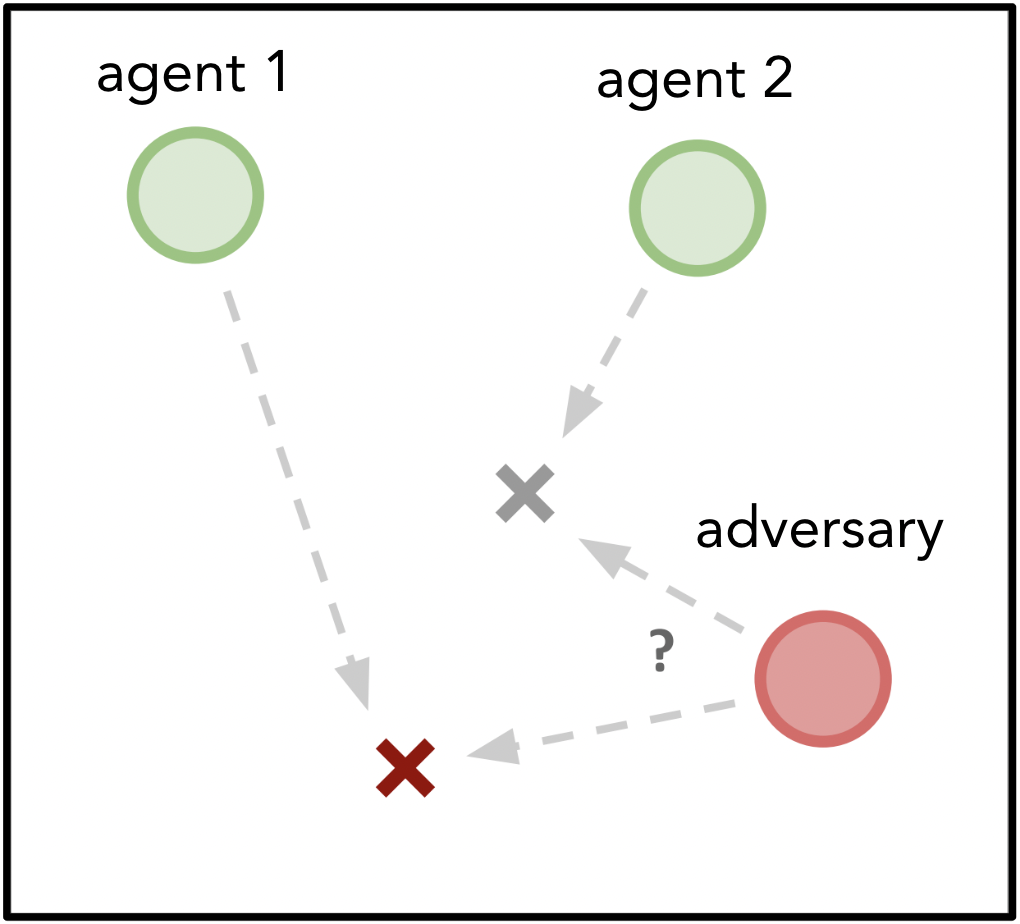}}
\caption{ParticleWorld physical deception environment.}
\end{center}
\vskip -0.2in
\end{figure}

We use a variant of the physical deception task described in \cite{particle-world}. This environment consists of $N$ landmarks, $N$ green ``good" agents and a single red adversary agent within a 2D world. 

In our variant, one of the landmarks is the ``target", but neither the good agents nor the adversary are initially told which one. The $N$ green agents receive a joint reward based on the minimum distance to the target landmark, with each agent's contribution weighted by a randomly generated reward coefficient $\eta_i \sim \mathrm{Uniform}[0, 1]$. Similarly, the adversary is penalized based on its distance from the target.

The episode ends either after a fixed time-limit, or when the adversary reaches any landmark. If this is the target landmark, the adversary receives a positive reward, otherwise a negative penalty (both time--scaled).
\begin{align}
   r_{good}(t) &= -\min_i  \left\{ d(\mathbf{x}_{i,t}, \mathbf{x}_{target}) \right\} \\\nonumber&+ d(\mathbf{x}_{adv}, \mathbf{x}_{target}) \\
   r_{adv}(t) &= -d(\mathbf{x}_{adv}, \mathbf{x}_{target})  \\\nonumber&+ \mathbb{I}[\mathbf{x}_{adv} = \mathbf{x}_{other}](1 - t/T) \\\nonumber&- \mathbb{I}[\mathbf{x}_{adv} = \mathbf{x}_{target}](1 - t/T)
\end{align}

where $d$ is Euclidean distance, $\mathbf{x}_{target}$ is the position of the target, $\mathbf{x}_{other}$ is the position of the non-target landmark, $\mathbf{x}_{i,t}$ is the position of good agent $i$ at time $t$, $\mathbf{x}_{adv,t}$ is the position of the adversary agent at time $t$, and $T$ is the maximum episode length.

The adversary is incentivized to find and navigate to the target as quickly as possible. On the other hand, the green agents are incentivized to keep the adversary uncertain as long as possible while accumulating reward.

\paragraph{Observations}
Each agent policy takes in a vector observation indicating the relative positions of landmarks and other agents. The good agents also can observe the weighted sum of their distances to the target landmark (weighted via their reward coefficients), whereas the adversary must rely on observing other agents' behavior to try and determine which landmark is the target.

\paragraph{Actions}
Each agent moves via a discrete action space.

\paragraph{Beliefs} In this scenario each agent is trained with two sets of first-order beliefs:
\begin{enumerate}
    \item Which landmark is the target?
    \item What are the reward coefficients for each agent?
\end{enumerate}







\subsection{Training}

We use Multi-Agent Proximal Policy Optimization (MAPPO) to train all agents in our experiments, under the paradigm of centralized training with decentralized execution (CTDE) \cite{yu2022mappo}. Our training procedure alternates between optimizing the policy for the good agents and the policy for the adversary, where one policy remains fixed and the weights other are trained; we swap every 100k timesteps.

\section{Preliminary Results}

\begin{figure}[htbp]
\begin{center}
\centerline{\includegraphics[width=0.8\linewidth]{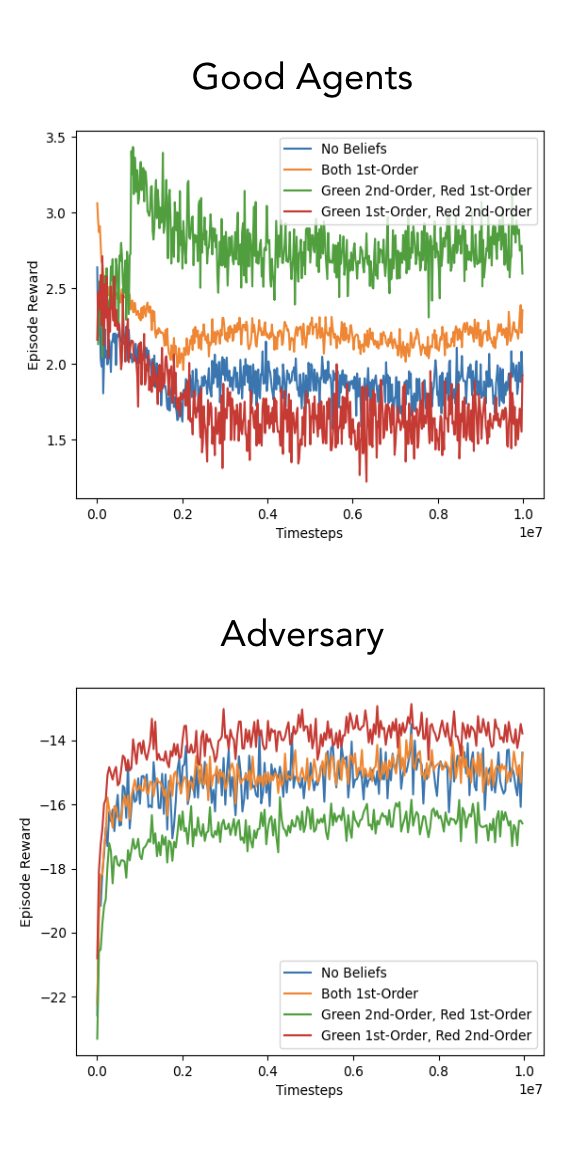}}
\caption{Training curves for agents on ParticleWorld physical deception task, in various belief configurations, averaged across 5 random seeds. Good agents trained with the 2nd-order belief intrinstic reward significantly outperform the other variants. Similarly, adversaries trained with the 2nd-order belief intrinsic perform better than their 1st-order belief and no-belief baseline counterparts.}
\label{fig:training-curve}
\end{center}
\end{figure}

We trained agents with various belief-prediction configurations on the physical deception task with $N=2$ landmarks; training curves are shown in Figure \ref{fig:training-curve}, and the mean episodic reward achieved by the final policies are shown in Table \ref{adversary-rewards}. We report the mean episode reward obtained with the best hyperparameter setting over 20 episodes, for each of 5 random seeds.

We find that agents with the 2nd-order intrinsic reward perform significantly better in relation to the opposition. This phenomenon is observed for both the green good agents and the red adversary.


\subsection{Qualitative Analysis of Observed Strategies}

We qualitatively assess and summarize the strategies observed with the final trained policies from each of the configurations we considered below.

\paragraph{Baseline (no beliefs)} Each green agent drifts towards a unique landmark. Red adversary appears to drifts randomly.

\paragraph{1st-order beliefs only (all agents)} Similar behavior to baseline.

\paragraph{2nd-order beliefs (green agents)} Each green agent drifts towards a specific landmark. In some episodes. green agents swap between landmarks.

\paragraph{2nd-order beliefs (red adversary)} Red tends to be more decisive, moving quickly to landmark.

In both cases we observe that the incorporation of the 2nd-order intrinsic reward tends to lead to the exhibition of more complex strategies that do not seem to be discovered with the baseline MARL approach, or even when learning with 1st-order beliefs alone.

\section{Ongoing and Future Work}
Although preliminary results indicate our approach may be effective, they are with respect to a single, relatively simple environment. We are currently examining more complex multi-agent tasks with more varied social dynamics, and additionally scaling the approach to scenarios with more (or even an arbitrary number of) agents.

Beyond continuing to experiment with other environments, we are particularly interested in studying the efficacy of our approach in communication; both in more traditional cooperative scenarios as well as potentially in competitive tasks.

We are also interested in a more thorough investigation of our concept-residual approach in comparison with the standard whitening or bottleneck approaches \cite{chen2020concept, koh2020concept}.


\section*{Acknowledgements}

This work is supported by the Defense Advanced Research Projects
Agency (DARPA) under Contract No. HR001120C0036, and by the AFRL/AFOSR award FA9550-18-1-0251.



\bibliography{example_paper}
\bibliographystyle{tom2023}




\end{document}